%% file: main.tex
\pdfoutput=1

\documentclass[11pt]{article}

\usepackage[final]{acl}

\usepackage{times}
\usepackage{latexsym}

\usepackage[T1]{fontenc}

\usepackage[utf8]{inputenc}

\usepackage{microtype}

\usepackage{inconsolata}

\usepackage{graphicx}

\usepackage{xcolor}
\usepackage{hyperref}
\usepackage{url}
\usepackage{tabularx}
\usepackage{array}
\usepackage[export]{adjustbox}
\usepackage{multirow, makecell}
\usepackage{booktabs}
\usepackage{varwidth}
\usepackage{xspace}
\usepackage{graphicx}
\usepackage{wrapfig}
\usepackage{pgfplots}
\pgfplotsset{compat=1.18}
\usepackage{tikz}
\usepackage{subcaption}
\usepackage{mdframed}
\usepackage{enumitem}
\usepackage{algorithm, algpseudocode}
\usepackage{amsthm}
\usepackage{amsmath}
\usepackage{amsfonts}
\usepackage{amssymb}
\usepackage{breqn}
\usepackage[symbol]{footmisc}
\usepackage{fontawesome5}

\theoremstyle{definition}
\newtheorem{definition}{Definition}[section]
\newtheorem{theorem}{Theorem}[section]

\newcommand\blfootnote[1]{%
  \begingroup
  \renewcommand\thefootnote{}\footnote{#1}%
  \addtocounter{footnote}{-1}%
  \endgroup
}

\title{Estimating Privacy Leakage of Augmented Contextual Knowledge \\ in Language Models 
}

 \author{James Flemings\textsuperscript{1}\thanks{Initial work conducted while interning at TikTok}\space\space Bo Jiang\textsuperscript{2}\space\space Wanrong Zhang\textsuperscript{2}\space\space Zafar Takhirov\textsuperscript{2}\space\space Murali Annavaram\textsuperscript{1} \\
         \textsuperscript{1}University of Southern California\space\space\textsuperscript{2}TikTok \\
         \texttt{\{jamesf17, annavara\}@usc.edu } \\
         \texttt{\{bjiang518, imwanrongz, z.tahirov\}@gmail.com}
         }

\begin{document}
\maketitle

\input{body/abstract}
\input{body/introduction}
\input{body/background}
\input{body/methodology}
\input{body/experiments}
\input{body/related_works}
\input{body/conclusion}
\input{body/limitations}
\input{body/acknowledgements}

\bibliography{main}

\newpage

\input{body/appendix}

\end{document}

%% file: body/abstract.tex
\begin{abstract}
    Language models (LMs) rely on their parametric knowledge augmented with relevant contextual knowledge for certain tasks, such as question answering. However, the contextual knowledge can contain private information that may be leaked when answering queries, and estimating this privacy leakage is not well understood. A straightforward approach of directly comparing an LM's output to the contexts can overestimate the privacy risk, since the LM's parametric knowledge might already contain the augmented contextual knowledge. To this end, we introduce \emph{context influence}, a metric that builds on differential privacy, a widely-adopted privacy notion, to estimate the privacy leakage of contextual knowledge during decoding. Our approach effectively measures how each subset of the context influences an LM's response while separating the specific parametric knowledge of the LM. Using our context influence metric, we demonstrate that context privacy leakage occurs when contextual knowledge is out of distribution with respect to parametric knowledge. Moreover, we experimentally demonstrate how context influence properly attributes the privacy leakage to augmented contexts, and we evaluate how factors-- such as model size, context size, generation position, etc.-- affect context privacy leakage. The practical implications of our results will inform practitioners of the privacy risk associated with augmented contextual knowledge.
    \blfootnote{\faGithub\quad\url{https://github.com/james-flemings/context_influence}}
\end{abstract}

%% file: body/introduction.tex
\section{Introduction}
\begin{figure}[t!]
   \centering
   \includegraphics[width=\columnwidth]{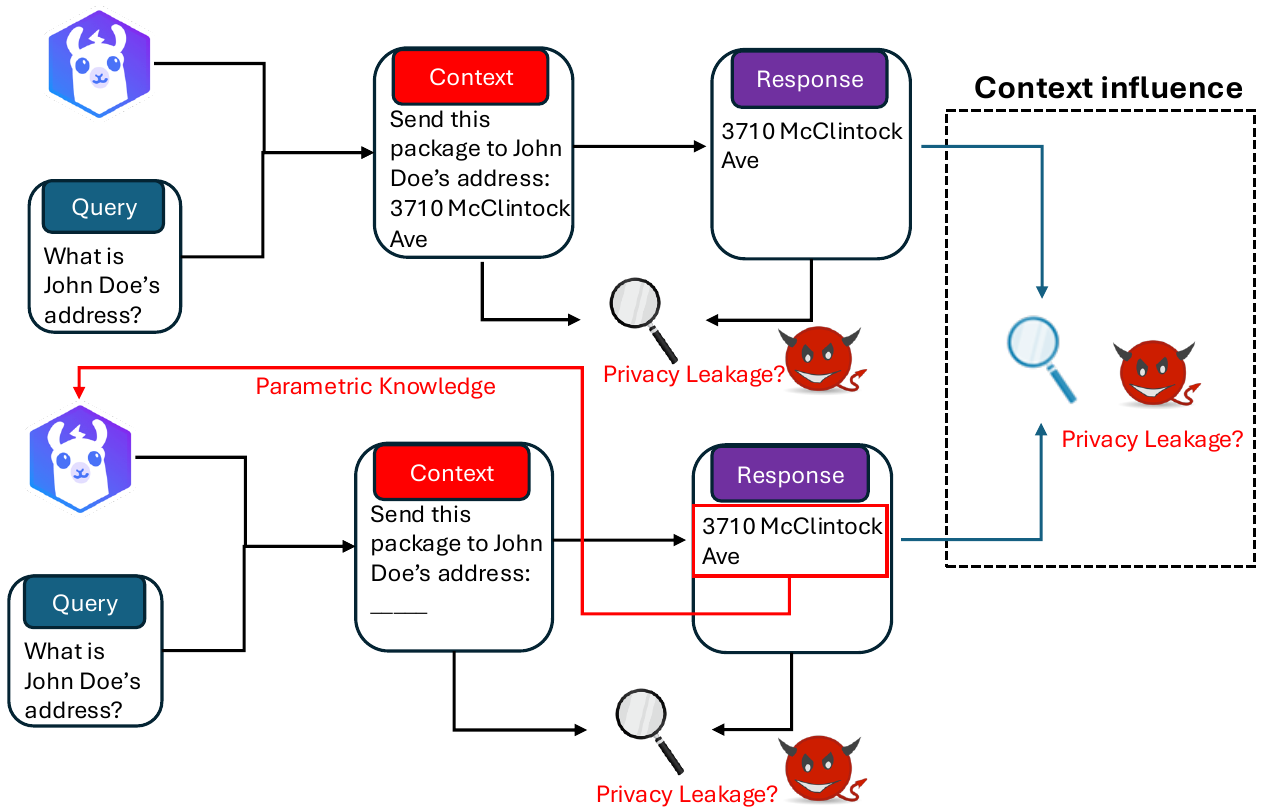}
   \caption{An illustration of properly measuring privacy leakage of contextual knowledge by comparing output distributions with and without sensitive information.}
   \label{fig:overview}
\end{figure}

Language Models (LMs) can rely on two sources of knowledge during generation: (1) \emph{parameteric knowledge}, which is information from the LM's pre-training corpora encoded within the model parameters \cite{devlin2018bert, radford2019language, petroni2019language}; (2) \emph{contextual knowledge}, which is additional information passed into the input prompt \cite{kwiatkowski2019natural, joshi2017triviaqa}. For certain downstream tasks, such as question-answering, it is essential to augment prompts containing a question/instruction with relevant context for LMs. However, a recent concern is that both the parametric and contextual knowledge may contain private information. Prior work has shown that privacy leakage of parametric knowledge often occurs from memorized pre-training data \cite{carlini2019secret}. On the other hand, we focus on the privacy leakage of augmented contexts, which can occur when an LM regurgitates them \cite{wang2023decodingtrust, priyanshu2023chatbots}. 

\input{figures/qual_n_gram}

Consider the example shown in Figure \ref{fig:overview}. An augmented context contains John Doe's address, and a user queries an LM asking for John Doe's address. If the output of the LM contains John Doe's address, then the straightforward approach of comparing the output against the augmented context would suggest there was privacy leakage from the context. This privacy evaluation was performed by prior works that studied data extraction attacks in RAG systems by prompting LMs to regurgitate the context \cite{zeng2024good, qi2024follow}. However, suppose we re-query the LM but remove (or mask out) John Doe's address from the context. If the LM still outputs John Doe's address, then surely the privacy leakage must derive from the LM's parametric knowledge. Hence, assuming that augmented contexts are not contained in the LM's parametric knowledge, which may not hold in practice \cite{golchin2023time, deng2023investigating, jiang2024investigating}, can overestimate the privacy leakage. Indeed, our results in Section \ref{sec:input_regurg} demonstrates this. Thus, accurately measuring context privacy leakage requires consideration of the existing parametric knowledge of the LM. 

Extensive research has examined how factors such as model size, prompt length, and training order contribute to the memorization and subsequent privacy leakage of an LM's parametric knowledge \cite{carlini2021extracting, carlini2022quantifying, biderman2024emergent, lesci2024causal}. However, there is a lack of understanding regarding the factors that cause privacy leakage of contextual knowledge. This is challenging as it involves separating the contributions of an LLM's parametric knowledge from the augmented context \cite{longpre2021entity, du2024context}, which has implications for solutions that adopt publicly pre-trained LMs to preserve privacy of contexts \cite{utpala2023locally, meisenbacher2024dp, flemings2024knowledge}.

These above-mentioned observations motivate the following fundamental research question: 
\begin{center}
    \vspace{-3pt}
    \emph{How can one estimate the privacy leakage of contextual information in a prompt given a specific parametric knowledge embedded in an LM?}
    \vspace{-3pt}
\end{center}
To answer this question, \textbf{we make the following contributions:} 
\setlist{nolistsep}
\begin{itemize}[leftmargin=*, itemsep=0em]
    \item We propose \emph{context influence}, a metric to principally quantify the privacy leakage of contextual information by measuring the output difference with and without a subset of the context, exemplified in Table \ref{tbl:qual_subset} with uni-grams. Context influence follows the analysis of differential privacy \cite{dwork2006differential}, a widely-adopted privacy notion.
    \item Then, using a slight reformulation of Context-aware Decoding \cite{shi2023trusting}, we show that context privacy leakage can be affected (1) explicitly when amplifying/deamplifying contextual knowledge during decoding, and (2) implicitly when contextual knowledge is out-of-distribution with respect to parametric knowledge. 
    \item Next, we experimentally show that our context influence metric properly attributes privacy leakage to the augmented contexts (Section \ref{sec:input_regurg}).
    \item Lastly, we experimentally evaluate how contextual and parametric knowledge, model capacity, context size, response position, and various context subsets affect context privacy leakage (Section \ref{sec:factors} \& \ref{sec:ith_n-gram}).
\end{itemize}

%% file: figures/qual_n_gram.tex
\begin{table*}[t]
\small
\begin{tabularx}{\textwidth}{p{0.15\textwidth} X}
\toprule
Current Generation & in a interview with Japanese defense minister, politician Antonio Inoki asked the defense minister about aliens and  \\ [4pt]
Next token & UFO  \\[4pt]
1-gram influence of context & \includegraphics[width=0.8\textwidth, valign=t]{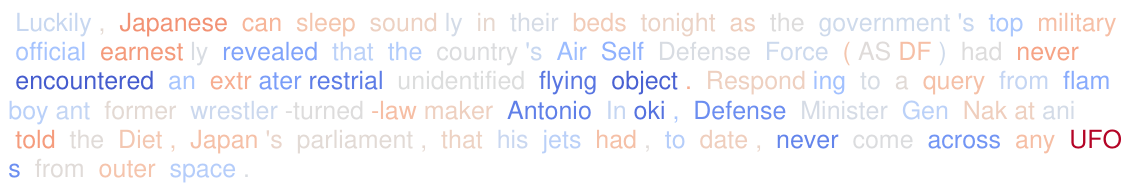} \\
& \includegraphics[width=0.4\textwidth]{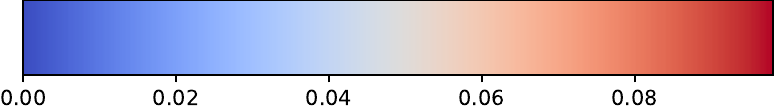} \\
\bottomrule
\end{tabularx}
\caption{A heatmap-like example of how \emph{context influence} measures privacy leakage of uni-gram tokens of a context from CNN-DM on the next token generation of LLaMA 3 8B. The next token generated is "UFO," and expectedly, the uni-gram with the highest leakage is "UFO." Interestingly, we see that words similar to "Japan" also strongly influenced LLaMA, while "flying" and  "object" did not.}

\label{tbl:qual_subset}
\end{table*}

%% file: body/background.tex
\section{Preliminaries}
Let $D=(d_1,...,d_N)$ be a list of tokens $d_i$, which we denote as a context. Let $p_\theta$ be an LM with model parameters $\theta$. We query $p_\theta$ with an instruction $\mathbf{x}$ for $D$ to generate a response $\mathbf{y}$. Specifically, we sample the response autogregressively from the likelihood probability distribution conditioned on the query $\mathbf{x}$, context $D$, and previously generated tokens $\mathbf{y}_{<t}$: $y_t\sim p_\theta(y_t|D,\mathbf{x},\mathbf{y}_{<t})$.

\subsection{Privacy}
To understand the privacy leakage of contextual information during decoding, we draw inspiration from Differential Privacy (DP) \cite{dwork2006differential, dwork2014algorithmic}, a strong privacy notion that gives a provable guarantee on the information leakage. We state the definition below.

\begin{definition}[Pure Differential Privacy (DP) \cite{feldman2021individual}] \label{def:dp}
    A randomized algorithm $\mathcal{A}$ satisfies $\epsilon$-DP if for all datasets $D=(d_1, ..., d_{N})$, it holds that
    \begin{align}
        \textstyle &\Pr[\mathcal{A}(D)\in E] \leq e^{\epsilon}\Pr[\mathcal{A}(D\!\setminus\!\{d_i\}) \in E], \text{and}\nonumber \\ 
        &\Pr[\mathcal{A}(D\!\setminus\!\{d_i\})\in E] \leq e^{\epsilon}\Pr[\mathcal{A}(D)\in E] \label{eq:pure_dp}
    \end{align} 
    for all $d_i\in D$ and all measurable sets $E$.
\end{definition}

This definition of DP follows the "add/remove" scheme where the neighboring datasets are defined by adding/removing one individual from the dataset. DP ensures that each individual in the dataset has at most $\epsilon$ information leakage. 

There are certain cases where the privacy losses of $\mathcal{A}$ (Eq. \ref{eq:pure_dp}) can vary substantially depending on the dataset $D$ and the realized output of the algorithm. Furthermore, the privacy loss bound $\epsilon$ is not informative about the privacy loss incurred to individuals $d_i$ in the dataset $D$. Hence, we define \textit{ex-post} per-instance DP that addresses these.

\begin{definition}[\textit{Ex-post} per-instance differential Privacy \cite{redberg2021privately}]\label{def:ex_post_DP}
    A randomized algorithm $\mathcal{A}$ satisfies $\epsilon(\cdot)$-\textit{ex-post} per-instance differential privacy for an individual $d_{i}$ and a fixed dataset $D$ at an outcome $\mathcal{A}(D)=o$ for $o\in\text{Range}(A)$ if 
    \begin{equation}\label{eq:ex-post_dp}
        \textstyle \left | \log \left (\frac{\Pr[\mathcal{A}(D)=o]}{\Pr[\mathcal{A}(D\setminus\{d_i\})=o]} \right  ) \right | \leq \epsilon(o, D, D\setminus\{d_{i}\}).
    \end{equation}
\end{definition}

\subsection{Context-aware Decoding}
Satisfying the DP definition requires controlling the amount of privacy that can be leaked by the context during generation. To this end, we borrow ideas from prior work that focused on amplifying contextual information by utilizing Pointwise Mutual Information (PMI) to measure the LM's dependence on the context, then applying this measurement to the decoding process to explicitly steer the LM's focus on the context \cite{van2022mutual, shi2023trusting}. The goal of these prior works does not involve privacy; rather, they are focused on reducing hallucinations by forcing the model to focus more on the context. However, these prior works provide the appropriate foundation to build on for estimating privacy from the context. PMI is defined as 
\begin{equation*}
    \textstyle \text{pmi}(p_\theta(y_t;D,\mathbf{x},\mathbf{y}_{<t}))=\log\left(\frac{p_\theta(y_t|D,\mathbf{x},\mathbf{y}_{<t})}{p_\theta(y_t|\mathbf{x},\mathbf{y}_{<t})}\right).
\end{equation*}
PMI measures the association of event $y_t$, predicting a specific token, and event $D$, the presence of context. The term $p_\theta(y_t|\mathbf{x}, \mathbf{y}_t)$ is the prior probability, representing the model's parametric knowledge $\theta$ without the context $D$, whereas the likelihood $p_\theta(y_t|D,\mathbf{x}, \mathbf{y}_t)$ represents the model's updated beliefs with the context $D$. To reduce LM hallucinations, one approach is to leverage PMI by multiplying a weighted PMI with the likelihood: 
\begin{align}
&y_t\sim\overline{p}_{\theta}( y_{t}|D,\mathbf{x},\mathbf{y}_{<t}) \propto \label{eq:cad} \\
&p_{\theta}(y_t|D,\mathbf{x},\mathbf{y}_{<t})\exp\left[\text{pmi}(p_{\theta}(y_t;D,\mathbf{x},\mathbf{y}_{<t}))\right]^{\beta} \nonumber
\end{align}
This formulation, known as \textbf{Context-aware Decoding (CAD)} \cite{shi2023trusting}, helps the LM focus on the context. 


%% file: body/methodology.tex
\section{Estimating Context Privacy}
\subsection{Motivation}
As we argued in the Introduction section, only comparing the LM's output to the augmented context is insufficient for measuring context privacy leakage. Alternatively, one could estimate the privacy risk by performing Membership Inference Attacks (MIAs) \cite{shokri2017membership, jagielski2020auditing}. However, this requires instantiating an attacker, which could severely underestimate the privacy leakage, and MIAs have been shown to be oftentimes ineffective on LMs \cite{duan2024membership}.

Instead, we take a different approach by utilizing the privacy analysis from DP, which provides a strong guarantee that bounds the privacy leakage to any dataset $D$, individual $d_i$, and output events $E$. In particular, we follow the observation that sampling-based decoding naturally satisfies the randomized output requirement of differential privacy with respect to the context \cite{flemings2024differentially}. Moreover, the neighboring definition of DP allows us to separate the contribution from parametric and contextual knowledge, by comparing the output probability distributions with and without subsets of the context. However, the $\epsilon$ bound from Eq. \ref{eq:pure_dp} does not provide much insights into the privacy risks of the augmented context, since $\epsilon$ is independent of the context. By using ex-post per-instance DP, we can directly calculate a privacy loss $\epsilon(o, D, D\setminus \{d_i\})$ that depends on important parameters-- such as the context $D$, neighboring contexts $D\setminus \{d_i\}$, generated token $o$-- which acts as a privacy auditing tool to analyze how these parameters affect the context privacy leakage. 

\subsection{Context Influence}
We now introduce \textit{context influence} below.

\begin{definition}[Context influence on next token]\label{def:mem}  Let $D$ be the context, and define $D_{i, n} = (d_{in}, ..., d_{(i+1)n})$ to be the $i$-th token $n$-gram of $D$. Let $\mathbf{x}$ be an input query, $p_\theta$ be an LM, and $\mathbf{y}_{<t}$ be the previous generations from $p_{\theta}$. Then we say that the context influence of $D_{i, n}$ on $p_\theta$ for an input query $\mathbf{x}$ when generating the next token $y_t$ is   
\begin{align}\label{eq:mem}
        &\tau_{i, n}(p_{\theta}, D, \mathbf{x}, \mathbf{y}_{<t}, y_t) = \\
        &| \underbrace{\log p_{\theta}(y_t|D,\mathbf{x},\mathbf{y}_{<t})}_{\substack{\text{output probability of $y_t$} \\ \text{given the context $D$}}} \!-\! \underbrace{\log p_{\theta}(y_t|D \!\setminus\! D_{i, n},\mathbf{x},\mathbf{y}_{<t})}_{\substack{\text{output probability of $y_t$ with } \\ \text{$D_{i, n}$ removed from context}}} | \nonumber
    \end{align}
\end{definition}
The use of $i$-th token $n$-grams generalizes the privacy level\footnote{We restrict the possible substrings to $n$-grams because they still produce natural text (contiguous substrings) while providing a granular measurement of the context influence.}, where $n=1$ roughly corresponds to word-level privacy \cite{xu2020differentially, feyisetan2020privacy} while $n=|D|$ corresponds to document-level privacy \cite{mattern2022limits, utpala2023locally} denoted as $D_{i, n} = D$. Rather than bounding the absolute value of the logs-odds ratio, as is done with DP, definition \ref{def:mem} directly uses this quantity as a lower-bound estimate of the privacy leakage of the $i$-th token $n$-gram $D_{i, n}$ for a fixed context $D$ when releasing the next token $y_t$. From an adversarial perspective, Definition \ref{def:mem} describes how confidently an attacker could infer whether the $i$-th token $n$-gram $D_{i, n}$ is part of the context $D$.

Furthermore, context influence measures how much the $i$-th token $n$-gram $D_{i, n}$ of the context $D$ influences the LM's prediction on the prompt data $(D, \mathbf{x}, \mathbf{y}_{<t})$. If $p_\theta$ is strongly influenced by $D_{i, n}$, then the removal of $D_{i, n}$ from the context $D$ would likely change the next token generation $y_t$ of $p_\theta$. Conversely, if the context influence is small, then that means the likelihood of generating $y_t$ with $p_\theta$ marginally changes with the removal of $D_{i, n}$ from the context. Thus, the next token from $p_\theta$ mostly depends on the remaining context $D\setminus D_{i, n}$, the current generation $\mathbf{y}_{<t}$, and its parametric knowledge $\theta$. 

To measure the total context influence over an entire generation $\mathbf{y}$, we simply sum the context influence for each generated token $y_t$ which is analogous to the basic composition property of DP \cite{dwork2014algorithmic}.
\begin{definition}[Context influence on response]\label{def:total_infl} We say that the context influence of $D_{i, n}$ on $p_\theta$ when generating the response $\mathbf{y}$ is the following:  
    \begin{equation*}
        \textstyle \tau_{i, n}(p_{\theta}, D, \mathbf{x}, \mathbf{y}) = \sum_{t} \tau_{i, n}(p_{\theta}, D, \mathbf{x}, \mathbf{y}_{<t}, y_t).
    \end{equation*}
\end{definition}
In our experimental evaluations, we are interested in measuring the expected context influence of each $i$-th token $n$-gram, regardless of the exact context and input query. We formalize this below:

\begin{definition}\label{def:gen_infl} \textbf{Expected context influence} on an LM $p_{\theta}$ is
    \begin{equation*}
        \tau_{i, n}(p_{\theta}) = \mathop{\mathbb{E}}_{(D, \mathbf{x})} \left[\tau_{i, n}(p_{\theta}, D, \mathbf{x}, \mathbf{y} \sim p_{\theta}(\mathbf{y} | D, \mathbf{x}))\right]
    \end{equation*}
\end{definition}
The equation from Definition \ref{def:gen_infl} can be directly estimated by using a set of pairs containing the context and its corresponding input query from the dataset $\mathcal{D}$. Each pair of context and input query $(D, \mathbf{x})\in\mathcal{D}$ is used to generate a response $\mathbf{y} \sim p_{\theta}(\mathbf{y} | D, \mathbf{x})$ from the LM. The resulting \textbf{estimator} is the following expression: 
\begin{align}
     \textstyle &\hat{\tau}_{i, n}(p_{\theta}) = \label{eq:estim_infl} \\
     &\frac{1}{|\mathcal{D}|} \sum_{(D, \mathbf{x})\in\mathcal{D}} \tau_{i, n}(p_{\theta}, D, \mathbf{x}, \mathbf{y} \sim p_{\theta}(\mathbf{y} | D, \mathbf{x})). \nonumber
\end{align}
And $\hat{\tau}_{|D|}(p_{\theta})$ denotes the special case in which we are measuring the influence of the entire context. 

One last technicality: context influence only works for sampling-based algorithms, such as temperature sampling \cite{ackley1985learning}, and not for greedy decoding algorithms, such as argmax. However, top-p \cite{holtzman2019curious} or top-k sampling \cite{fan2018hierarchical} can cause potential errors in the context influence calculation unless the selected indices are equal for both $D$ and $D\setminus D_{i, n}$. For the remainder of our work, we focus only on temperature sampling. Specifically, we generate the responses $\mathbf{y}$ to be used for measuring context influence by using a slight reformulation of CAD (Eq. \ref{eq:cad}), giving us granular control over how much of the contextual knowledge is used during decoding. We introduce \textbf{Context Influence Decoding (CID)} below:

\begin{definition}\textbf{CID} samples from is a linear interpolation between the likelihood and the prior logits using a weighing term $0\leq \lambda < \infty$
\begin{align}\label{eq:CID_dist}
    &\overline{p}_{\theta, \lambda}(y_t|D,\mathbf{x},\mathbf{y}_{<t})= 
    \sigma[(\lambda\text{logit}_\theta(y_t|D,\mathbf{x},\mathbf{y}_{<t})\nonumber\\ 
    &+(1-\lambda)\text{logit}_\theta(y_t|\mathbf{x},\mathbf{y}_{<t}))/T]
\end{align}
\end{definition}
where $\sigma$ is the softmax function and $T$ is the temperature parameter where $T>1$ resulting in a more uniform distribution (i.e. higher entropy) and $0<T<1$ forcing a sharper output distribution. 

Note that CID reformulates CAD by utilizing a tunable parameter $\lambda$ that explicitly controls the influence level of a context during decoding. We start with the prior logits $\text{logit}_\theta(y_t|\mathbf{x},\mathbf{y}_{<t})$, which contains no information about the context $D$, and increasingly adds more information from the PMI, which can leak information about the context $D$, by increasing the weighing parameter $\lambda$. 

\subsection{Privacy Leakage with Context Influence }\label{sec:cid_dp}
Because by definition context influence measures the privacy loss of individual $n$-grams, CID naturally achieves $n$-gram level $\tau_{i, n}(p_{\theta, \lambda}, D, \mathbf{x}, \mathbf{y}_{<t})$-\textit{ex-post} per-instance DP (Definition \ref{def:ex_post_DP}) by using our context influence definition to bound the $i$-th token $n$-gram privacy loss. However, it is possible to achieve the stronger $n$-gram level $\epsilon$-DP definition by essentially selecting $\textstyle \lambda^*$ such that generating the next token by CID satisfies 
\begin{equation}\label{eq:dp_cid}
    \max_{D, i, y_t} \tau_{i, n}(\overline{p}_{\theta, \lambda^*}, D, \mathbf{x}, \mathbf{y}_{<t}, y_t) \leq \epsilon
\end{equation}
for a fixed $n$. The proof can be found in Appendix \ref{sec:add_dp}. Meaning regardless of the context $D$, the previously generated tokens $\mathbf{y}_{<t}$, and the next token $y_t$, the $i$-th token $n$-gram $D_{i, n}$ influences the LM $\theta$ by at most $\epsilon$; hence, DP bounds the privacy leakage with a context-independent value. For our work, we want to explicitly measure how the privacy leakage changes with respect to the aforementioned variables, hence why chose the analysis of \textit{ex-post} per-instance DP. This gives a guarantee that the privacy leakage when releasing $y_t$ is at least $\tau_{i, n}(\overline{p}_{\theta, \lambda}, D, \mathbf{x}, \mathbf{y}_{<t}, y_t)$, which follows the more practical direction of privacy auditing \cite{jagielski2020auditing}.

Lastly, to better understand which factors affect context privacy leakage, we will use CID to connect context influence directly with PMI.
\begin{theorem}\label{thm:priv_loss_pmi}
    Let $\lambda \geq 0$. Then, the influence of $D_{i, n}$ on the response $y_t$ generated from CID is
    \begin{align}
        \textstyle &\tau_{i, n}(\overline{p}_{\theta, \lambda}, D, \mathbf{x}, \mathbf{y}_{<t}, y_t) \! \propto \! \lambda | \text{pmi}(p_{\theta}(y_t;D,\mathbf{x},\mathbf{y}_{<t}) \nonumber \\
        &)- \text{pmi}(p_{\theta}(y_t;D \! \setminus \! D_{i, n},\mathbf{x},\mathbf{y}_{<t})) |
    \end{align}
\end{theorem}
\textit{Proof.} We defer the proof to Appendix \ref{sec:proof}. \qed

Theorem \ref{thm:priv_loss_pmi} reveals that the privacy leakage of the $i$-th token $n$-gram of the context depends on two key factors: (1) the difference in PMIs, which quantifies how much the generated next token relies on the $i$-th token $n$-gram of the context, and (2) the parameter $\lambda$, which directly controls the influence of context knowledge on the CID. Hence, the context privacy leakage can be exacerbated in two scenarios:
\begin{itemize}[leftmargin=*, nosep]
    \item The context $D$ is out-of-distribution with respect to the LM's parametric knowledge $\theta$, and the subset of contextual information we are interested in is sufficiently large (i.e. large $n$), both maximizing the difference of PMIs. Various factors-- including the type of contextual $(D, \mathbf{x})$ and parametric knowledge $\theta$, model size $|\theta|$, and context size $|D|$-- can cause the contextual and parametric knowledge to diverge, which we experimentally analyze in Section \ref{sec:factors}. Additionally, in Section \ref{sec:ith_n-gram}, we compare how different subsets of contextual knowledge influence an LM. 
    \item When the contextual knowledge is amplified (higher $\lambda$) to reduce context-conflicting hallucination. In Section \ref{sec:input_regurg} and \ref{sec:factors}, we quantify how changes in $\lambda$ lead to higher privacy risks.
\end{itemize}

%% file: body/experiments.tex
\input{figures/input_regurgitation_table}

\section{Experimental Evaluations}
\subsection{Experimental Setup}
\textbf{Datasets.} We perform our experimental evaluations on two open-ended generation tasks: \emph{summarization via CNN-DM} \cite{see2017get}, a collection of English news articles written by journalists at CNN and the Daily Mail, and \emph{long-form question-answering via PubMedQA} \cite{jin2019pubmedqa}, a dataset from the biomedical domain and contexts available. Appendix \ref{sec:exam_promp} contains example prompts used for context influence. Each context document is truncated by $2048$ and $1024$ for PubMedQA and CNN-DM, respectively. 

\textbf{Metrics.} We evaluate the generation quality along two dimensions: \textit{similarity} and \textit{faithfulness}. For similarity, we employed F1 ROUGE-L \cite{lin2004rouge} and F1 BERTScore \cite{zhang2019bertscore} to measure lexical and semantic similarity between the response and the reference, respectively. For faithfulness, we used FactKB \cite{feng2023factkb} to measure the faithfulness of the response to the context. Our calculation of context influence uses the empirical estimator in Eq. \ref{eq:estim_infl} with CID using sampled contexts from CNN-DM and PubMedQA.

\textbf{Models.} Since context influence requires access to the entire model output distribution, we use open-source models due to closed-source ones restricting the output logits. We used OPT 1.3B \cite{zhang2022opt}, GPT-Neo 1.3B \cite{black2021gpt}, LLaMA 3 8B and LLaMA 3 8B IT (Instruct) \cite{dubey2024llama}, and Gemma 2 9B (Instruct) \cite{team2024gemma}. We set temperature parameter $T=0.8$, the response length to at most 50 tokens, and the number of responses for each dataset is $N=1000$. 

\subsection{Context Influence on Input Regurgitation} \label{sec:input_regurg}
First, we demonstrate how our context influence metrics offer improvements over directly comparing LLM output with augmented contexts. Following the untargeted attack evaluations by \citet{zeng2024good} we report:
\begin{itemize}[leftmargin=*, nosep]
    \item \textbf{Repeat Prompts:} The number of prompts yielding a response with at least half direct tokens from the context.
    \item \textbf{Rouge Prompts:} The number of prompts generating responses with a ROUGE-L score over 0.5. 
\end{itemize}
Next, we show that Repeat Prompts and Rouge Prompts can erroneously indicate privacy leakage from augmented PubMedQA abstracts if the LLM's parametric knowledge already includes PubMed abstracts. For this, we compare OPT 1.3B and GPT-Neo 1.3B, models with the same number of parameters and similarly follow the GPT-3 architecture \cite{brown2020language}. GPT-Neo was pre-trained on The Pile dataset \cite{gao2020pile}, which contains PubMed abstracts. In contrast, OPT was trained on a subset of The Pile that excludes PubMed abstracts \cite{zhang2022opt}. Therefore, we expect OPT 1.3B to show greater context privacy leakage, as it likely lacks PubMed abstracts in its parametric knowledge and must rely more on the augmented PubMedQA contexts.

Table \ref{tbl:input_regurg} displays the results. We observe that context influence accurately follows our expectation by correctly attributing the privacy leakage to the PubMedQA abstracts. However, we observe that both Repeat Prompts and Rouge Prompts are larger for GPT-Neo 1.3B than OPT 1.3B. Consequently, this suggests that LLMs are likely to leak the augmented contexts if they were trained on them. However, we argue that this privacy leakage should not be entirely attributed to the augmented context, but rather should be shared with the LLM's parametric knowledge, as is done in context influence. 

\subsection{Factors Contributing to Context Influence}\label{sec:factors}
\input{figures/main_results_table}

Next, we experimentally analyze the identified factors from Section \ref{sec:cid_dp} that could cause an LM to unintentionally leak contextual information.

\textbf{Context influence level $\lambda$.} First, we vary the context influence level $\lambda\in\{0.5, 1.0, 1.5\}$ for CID to see how it affects the measured context influence. From Table \ref{tbl:main}, we observe that for LLaMA 3 on CNN-DM, amplifying the context by increasing the influence level from $\lambda=1.0$ to $\lambda=1.5$ leads to a 10\% increase in F1 ROUGE-L due to 50\% more influence by the context. However, Table \ref{tbl:input_regurg} shows that this increased context influence is attributable to 50\% more input regurgitation, raising a key concern that amplifying contextual knowledge during decoding can lead to increased privacy risks. When we reduce the context influence level to $\lambda=0.5$, we observe that the context influence is reduced by 2.2x, leading to near-zero regurgitation of context. However, this comes at a cost of substantial utility degradation. Hence, completely reducing input regurgitation has a deleterious outcome on the utility. Additional influence levels $\lambda$ can be found in Appendix \ref{sec:add_exp_results}.

\textbf{Contextual knowledge ($D, \mathbf{x})$.} Next, we investigate how the type of context and instruction $D, \mathbf{x}$ affects context influence. From Table \ref{tbl:input_regurg} and Table \ref{tbl:main}, we observe that LMs performing abstractive summarization (CNN-DM) rely on/repeat the context more than long-form question-answering (PubMedQA). This means the type of contextual information and instruction have a substantial effect on context privacy. In particular, the query from CNN-DM explicitly instructs the LM to shorten the context, whereas for PubMedQA the LM could decide not to use the context to answer the query. Hence, one way to preserve context privacy would be to instruct the LM to utilize their parametric knowledge while reducing context hallucination. 

\textbf{Parametric knowledge $\theta$.} We compare context influence on multiple different parametric knowledge sources. Our results indicate that the choice of parametric knowledge can have a substantial affect on the context influence. In particular, we saw from Section \ref{sec:input_regurg} how just the inclusion of PubMed data in the pre-training data can effectively decrease the context influence of PubMedQA abstracts, creating a false sense of security that GPT-Neo preserves the privacy of the PubMedQA abstracts better than OPT. Hence, this raises an important caveat that smaller context influence does not necessarily imply better context privacy, as one must consider the public data used for pre-training \cite{tramer2022considerations}.

\textbf{Pre-trained vs fine-tuned.} From Table \ref{tbl:main}, we observe that LLaMA 3 IT is substantially influenced by the context more than just pre-trained LLaMA 3. This is intuitive as LLaMA 3 IT received further training in the form of supervised fine-tuning (SFT) and reinforcement learning with human feedback (RLHF) to align better with prompt answering. These additional steps, SFT and RLHF, help the model utilize the context more when answering queries, hence, increasing the context influence. Thus, the increased performance from fine-tuning results in larger context privacy leakage.

\textbf{Model size $|\theta|$.} We analyze the effect of model size $|\theta|$ on context influence for CID with regular decoding ($\lambda=1.0$). We used various sizes-- 125M, 350M, 1.3B, 2.7B, 13B, 30B, and 66B-- of OPT evaluated on PubMedQA. The results shown in Figure \ref{fig:model_size} depict a trend with some variability, but it generally shows that larger models are less influenced by the context. We hypothesize that larger models have a larger capacity to memorize their pre-training data, so they can rely on their parametric knowledge more than smaller models. 

\input{figures/analysis_plots}

\textbf{Context size $|D|$.} Additionally, we measured the effect of the context size $|D|$ on context influence for CID using OPT-1.3B. In this setup, we restrict the model to only the first $|D|$ tokens of context for generation and calculating context influence. Shown in Figure \ref{fig:context_size}, we observe that when the context is extremely small ($\leq 32)$, then the LM is substantially less influenced by the context. The context may not contain enough relevant information to help the model, and hence, it must rely on its parametric knowledge. However, as we increase the context size from 32 to 256, the model becomes more influenced by the context. After $|D| \geq 256$, the model maintains a relatively constant level of context influence. Hence, truncating the context has marginal affect on context privacy unless the size of the context is substantially reduced.

\textbf{Response position $y_t$.} Lastly, we measured how far along the prior generation (the size of $\mathbf{y}_{<t})$ affects how much OPT-1.3B is influenced by the context when generating the next token. More precisely, we measure the average context influence of the next token at the $t$-th position $\tau_{|D|}(p_{\theta}, D, \mathbf{x}, \mathbf{y}_{<t}, y_t)$ over all generations. As shown in Figure \ref{fig:gen_len_analysis}, we observe that the first 10 generated tokens by the model are influenced by the context the most. This is intuitive as the initial response generated by the model is small and nascent; hence, it must rely on the context more for the next token generations. But as the generated response size increases $|\mathbf{y}_t|$, the model can rely more on its parametric knowledge $\theta$ and the current generated response $\mathbf{y}_{<t}$ for generating the next token $y_t$. Thus, one can design privacy-preserving solutions that adopt an adaptive privacy level, where the privacy level is strict during the beginning of generating tokens, then is relaxed as more tokens are generated.

\subsection{Token $n$-gram Influence of Context}\label{sec:ith_n-gram}
In this section, we analyze the context influence of each $i$-th token $n$-gram in the context $\hat{\tau}_{i, n}(\overline{p}_{\theta, \lambda})$, i.e., we compare the output probability with and without the $i$-th token $n$-gram from the context to measure the influence. Due to the possibly large number of token $n$-grams, we only evaluate 100 contexts. Figure \ref{fig:n_gram_loc_analysis} shows the results for various token (128, 32, 8, 4)-gram influence on PubMedQA for OPT-1.3B with $\lambda=1.0$. We observe two trends: (1) Larger $n$-grams have higher peak context influence, which is intuitive given that the more information (larger $n$) is removed from the context, the more likely the output of the LM will change; (2) for seemingly all $n$, the context influence peaks for earlier $i$-th token $n$-grams, i.e. small $i$, then gradually declines for later $i$-th token $n$-grams. The results suggest that the model is influenced by information located earlier in the context than those located late, which might stem from the larger issue of position bias \cite{liu2024lost}. This implies that practitioners who want to control the influence of certain sequences can place privacy-sensitive ones toward the end of the context.

Next, we look at the context influence of each $i$-th token 128-gram for various OPT sizes, 125M, 350M, 1.3B, 6.7B, and 13B, in Figure \ref{fig:model_size_n_gram}. We observe that regardless of model size, the context influence peaks at the earlier token 128-grams, then gradually decreases for later ones. Generally, the context influence for most token 128-gram is strongest for OPT 13B and the lowest for OPT 6.7b, but this trend reverses towards the later 128-grams. Interestingly, we observe sporadic spikes in context influence for small LMs, OPT 125M and 350M, on the later token $128$-grams, suggesting that smaller LMs need to rely on more parts of the context.

\input{figures/n_gram_analysis}

%% file: figures/input_regurgitation_table.tex
\begin{table*}[!ht]
\centering
\small
\begin{tabular}{c c| c c c | c c c }
\toprule

& & \multicolumn{3}{c}{\textbf{CNN-DM}} & \multicolumn{3}{c}{\textbf{PubMedQA}} \\
\cmidrule(lr){3-5}
\cmidrule(lr){6-8}

\textbf{Model} & \textbf{Decoding $\lambda$} & $\hat{\tau}_{|D|}(\overline{p}_{\theta, \lambda})$ & \makecell{Repeat \\ Prompts} & \makecell{Rouge \\ Prompts} & $\hat{\tau}_{|D|}(\overline{p}_{\theta, \lambda})$ & \makecell{Repeat \\ Prompts} & \makecell{Rouge \\ Prompts} \\

\midrule

\multirow{3}{*}{LLaMA 3 8B} & 0.5 & 15.97 & 8 & 109 & 16.69 & 0 & 128\\
& 1.0 & 64.61 & 285 & 632 & 37.01 & 58 & 439 \\
& 1.5 & 98.99 & 429 & 882 & 70.91 & 123 & 669\\
\midrule

\multirow{3}{*}{OPT 1.3B} & 0.5 & 17.50 & 1 & 87 & 13.20 & 0 & 54\\
& 1.0 & 85.23 & 373 & 644 & 45.66 & 47 & 251\\
& 1.5 & 140.0 & 559 & 836 & 97.95 & 151 & 494\\

\midrule

\multirow{3}{*}{GPT-Neo 1.3B} & 0.5 & 15.16 & 5 & 87 & 11.20 & 0 & 53\\
& 1.0 & 77.87 & 338 & 571 & 38.79 & 54 & 268\\
& 1.5 & 130.47 & 637 & 822 & 77.91 & 206 & 622\\

\bottomrule
\end{tabular}
\caption{Measuring Context influence and input regurgitation of various influence levels $\lambda$.}
\label{tbl:input_regurg}
\end{table*}

%% file: figures/main_results_table.tex
\begin{table*}[th!]
\small
\centering
\resizebox{\linewidth}{!}{%
\begin{tabular}{lc|cccc|cccc}
\toprule

& & \multicolumn{4}{c}{\textbf{PubMedQA}} & \multicolumn{4}{c}{\textbf{CNN-DM}} \\
\cmidrule(lr){3-6}
\cmidrule(lr){7-10}
\textbf{Model} & \textbf{Decoding $\lambda$} & $\hat{\tau}_{|D|}(\overline{p}_{\theta, \lambda})$ & ROUGE-L & BERTS & FactKB & $\hat{\tau}_{|D|}(p_{\theta, \lambda})$ & ROUGE-L & BERTS & FactKB \\ 
\midrule

\multirow{3}{*}{OPT 1.3B} & 0.5 & 13.20 & 15.41 & 72.13 & 31.40 & 17.50 & 9.73 & 68.06 & 75.28 \\
& 1.0 & 45.66 & 16.51 & 72.81 & 37.38 & 85.23 & 16.84 & 72.09 & 88.24 \\
& 1.5 & 97.95 & 16.96 & 72.88 & 48.81 & 140.0 & 18.82 & 72.88 & 89.22 \\
\midrule

\multirow{3}{*}{GPT-Neo 1.3B} & 0.5 & 11.20 & 16.26 & 72.32 & 35.66 & 15.16 & 9.73 & 68.06 & 75.28 \\
& 1.0 & 38.79 & 18.47 & 73.65 & 52.36 & 77.87 & 15.97 & 71.54 & 93.66 \\
& 1.5 & 77.91 & 18.91 & 74.08 & 68.54 & 130.47 & 18.17 & 72.66 & 92.90 \\
\midrule

\multirow{3}{*}{LLaMA 3 8B} & 0.5 & 16.69 & 17.73 & 73.33 & 44.71 & 15.97 & 10.34 & 68.06 & 69.18 \\
& 1.0 & 37.01 & 19.20 & 74.66 & 49.63 & 64.61 & 17.42 & 72.17 & 85.60 \\
& 1.5 & 70.91 & 18.79 & 74.41 & 56.76 & 98.99 & 19.22 & 72.89 & 87.86 \\
\midrule

\multirow{3}{*}{LLaMA 3 8B IT} & 0.5 & 17.26 & 20.17 & 74.51 & 51.64 & 35.0 & 15.18 & 71.89 & 87.22 \\
& 1.0 & 66.39 & 21.47 & 75.47 & 56.64 & 92.25 & 22.53 & 73.35 & 98.26 \\
& 1.5 & 115.78 & 20.88 & 75.21 & 63.08 & 134.23 & 23.53 & 75.44 & 97.95 \\
\midrule

\multirow{3}{*}{Gemma 2 9B IT} & 0.5 & 26.68 & 18.52 & 74.03 & 35.60 & 41.76 & 14.49 & 71.73 & 87.56 \\
& 1.0 & 70.10 & 20.05 & 74.97 & 41.60 & 93.17 & 21.18 & 75.09 & 96.3 \\
& 1.5 & 111.07 & 18.52 & 74.03 & 35.60 & 149.33 & 21.60 & 75.22 & 96.31 \\
\bottomrule
\end{tabular}}
\caption{The context influence-hallucination tradeoff of different context influence levels of CID.}
\label{tbl:main}
\end{table*}

%% file: figures/analysis_plots.tex
\begin{figure*}[t!]
    \centering
    \begin{subfigure}[t]{0.325\linewidth}
        \centering
        \includegraphics[width=1.0\linewidth]{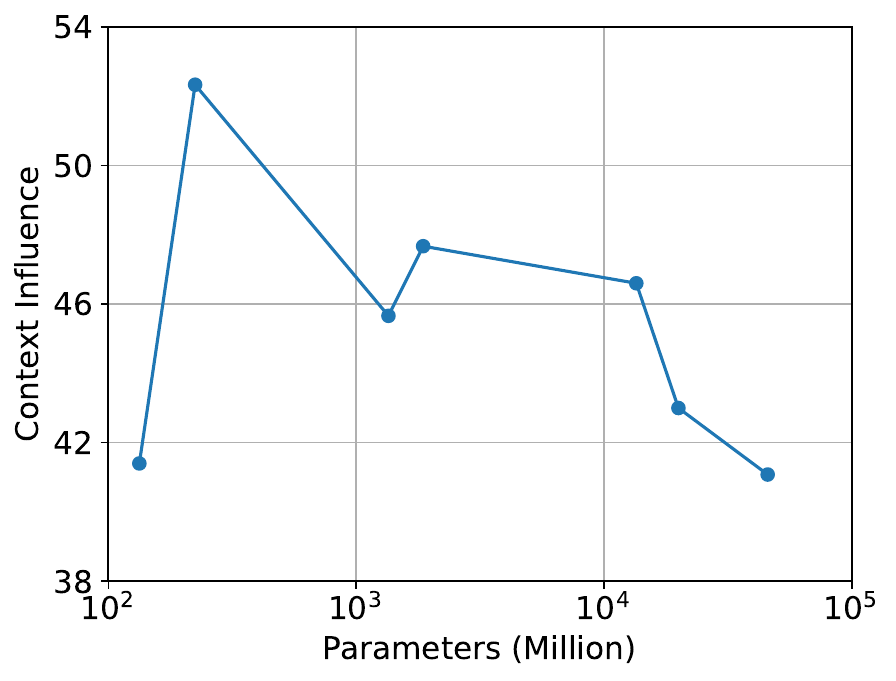}
        \caption{}
        \label{fig:model_size}
    \end{subfigure}
    \hfill 
    \begin{subfigure}[t]{0.32\linewidth}
        \centering
        \includegraphics[width=1.0\linewidth]{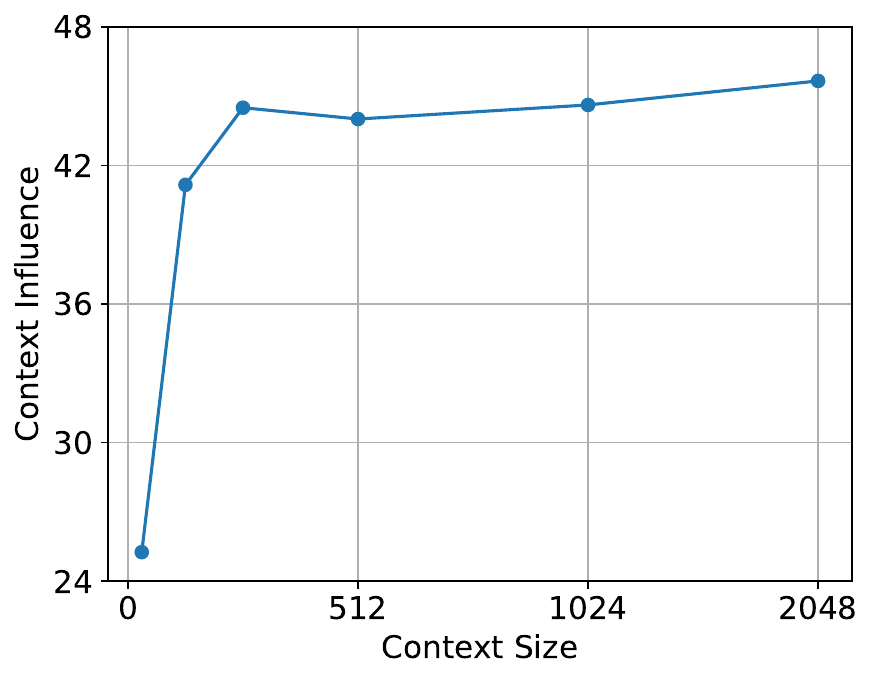}
        \caption{}
        \label{fig:context_size}
    \end{subfigure}
    \hfill 
    \begin{subfigure}[t]{0.305\textwidth}
        \centering 
        \includegraphics[width=1.0\linewidth]{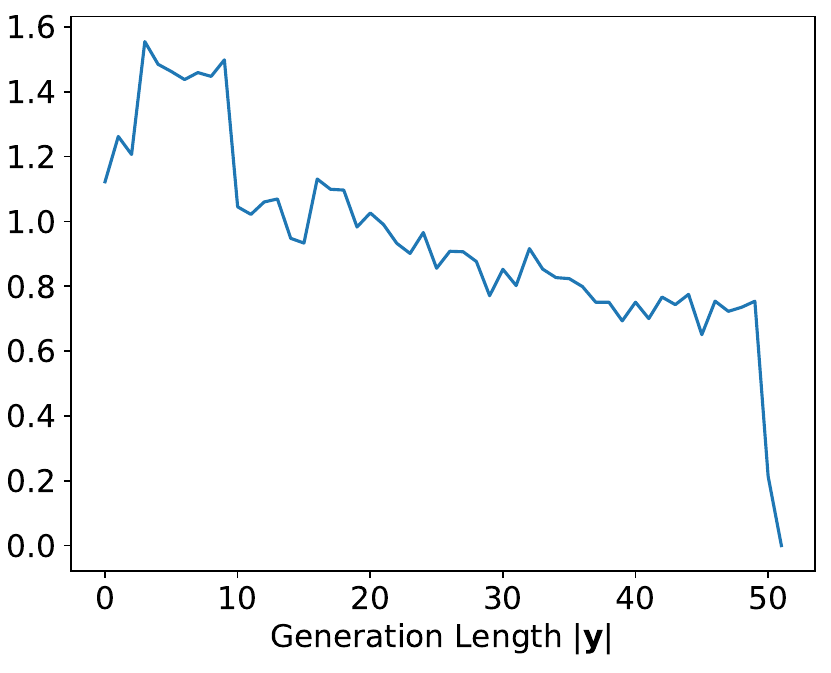}
        \caption{}
        \label{fig:gen_len_analysis}
    \end{subfigure}
    \caption{Measuring the affect of \textbf{(a)} model size, \textbf{(b)} context size, and \textbf{(c)} response size on context influence.}
    \label{fig:ablat}
\end{figure*}

%% file: figures/n_gram_analysis.tex
\begin{figure*}[th!]
    \centering
    \begin{minipage}[t]{0.45\textwidth}
        \centering 
        \includegraphics[width=1.0\linewidth]{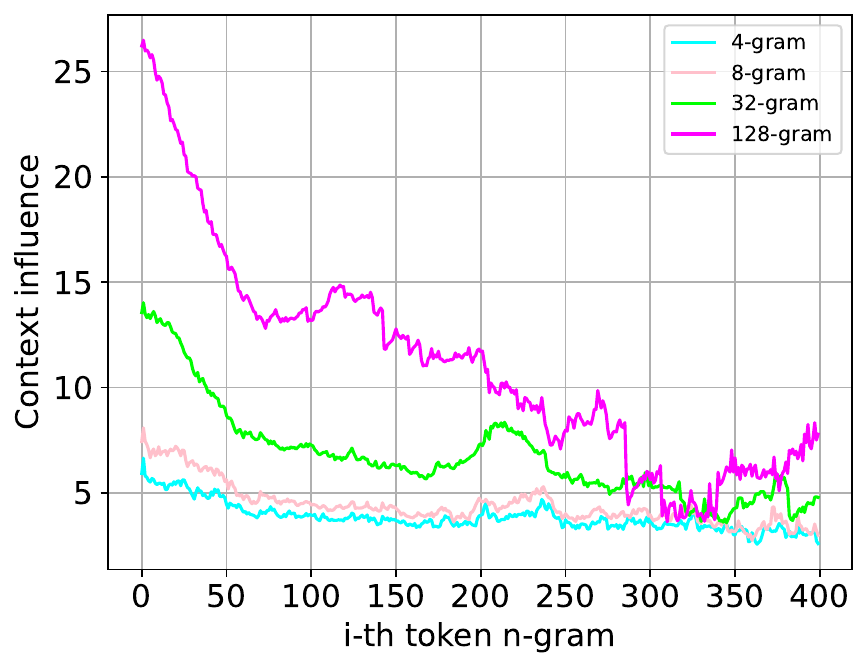}
        \captionof{figure}{Measuring token $n$-gram context influence of various $n$-grams on OPT 1.3B. }
        \label{fig:n_gram_loc_analysis}
    \end{minipage}\hfill
    \begin{minipage}[t]{0.45\textwidth}
        \centering 
        \includegraphics[width=1.0\linewidth]{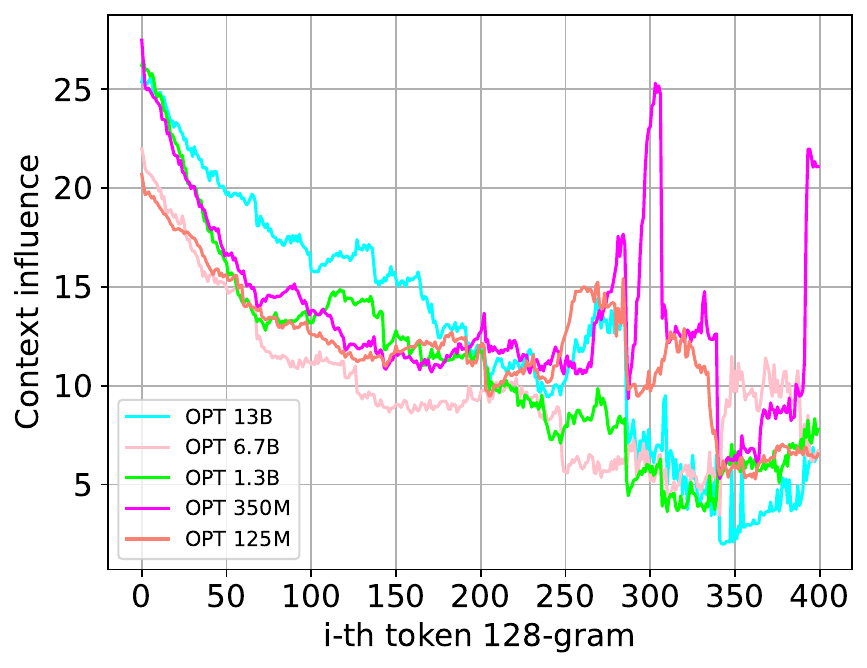}
        \captionof{figure}{Measuring the context influence of each $i$-th token $128$-gram for various sizes of OPT models.}
        \label{fig:model_size_n_gram}
    \end{minipage}
\end{figure*}

%% file: body/related_works.tex
\section{Related Works}
\textbf{Parametric Knowledge Leakage.} It has been demonstrated that inadvertent memorization of pre-training data can lead to privacy leakage \cite{carlini2019secret, song2019auditing} in the form of extraction attacks \cite{carlini2021extracting, thomas2020investigating}. Hence, there is extensive research on understanding the memorization dynamics of LMs \cite{tirumala2022memorization, zhang2023counterfactual, lesci2024causal, biderman2024emergent}, where it has been shown that various factors such as model size, data duplication, and prompt length increase memorization. From these results, works have proposed dataset curation techniques, such as data deduplication \cite{kandpal2022deduplicating}, to mitigate training data privacy leakage. In this work, we seek to conduct a similar analysis for augmented contexts.

\textbf{Contextual Knowledge Leakage.} Recent works have demonstrated that LMs can leak privacy-sensitive information provided to a prompt during inference via prompt regurgitation \cite{wang2023decodingtrust, priyanshu2023chatbots}. In particular, recent works have shown through the lens of contextual integrity theory \cite{nissenbaum2009privacy} that LMs lack the ability to effectively reason about the information sensitivity of contextual knowledge \cite{mireshghallah2024can, bagdasarian2024airgapagent, shao2024privacylens}. On the other hand, our analysis operationalizes \textit{exp-post} per-instance DP to understand the factors that unintentionally influence contextual knowledge leakage. \citet{zeng2024good, qi2024follow} investigated attacks on RAG systems that extract contextual knowledge, a similar setup and goal to our work. However, these works focus primarily on data extraction and implicitly assume that the retrieved contexts from the RAG database are not contained in the LM's parametric knowledge, which overly-attributes the privacy leakage to the contexts. Also related, \citet{huang2023privacy} investigated the privacy leakage of retrieval-based LMs, such as kNNs. Lastly, another body of work investigated MIAs for augmented contexts \cite{anderson2024my, wang2024membership, li2025generating}. Context influence can be viewed as inferring membership of a context.

\textbf{Context hallucination.} Our work follows prior work on summarization factuality where the response from an LM conflicts with an augmented context \cite{maynez2020faithfulness, pagnoni2021understanding}. We focus on hallucination mitigation during inference by utilizing PMI to amplify focus on contextual rather than parametric knowledge \cite{van2022mutual, shi2023trusting}. Our results demonstrate how these decoding methods affect the privacy leakage of contextual knowledge. Another body of work \cite{fernandes2021does, sarti2023quantifying, cohen2024contextcite, du2024context} measured an LM's reliance on an augmented context; however, the goal of these works is not motivated by privacy, and hence, their results/discussion are orthogonal to ours. 

%% file: body/conclusion.tex
\section{Conclusion}
Studying the influence of augmented context on the generations of LMs has crucial implications for privacy. Hence, our goal is to principally undertake this study to inform practitioners of the context privacy risks and design solutions with these results in mind. We introduced a principled definition for context influence to measure the privacy leakage of contextual knowledge. Then we measured context influence on various LMs for two types of open-ended generation tasks. We found that the choice of contextual and parametric knowledge, model capacity, context and response size, and token n-grams largely affect the privacy of contextual information. 

%% file: body/limitations.tex
\section{Limitations}
We defined context influence in a way that allowed us to connect it with pointwise mutual information and differential privacy. However, a limitation of this formulation is that it does not consider the entropy of the model during decoding. For example, the context influence of a more confident model will be smaller than that of a less confident one. One way to overcome this limitation is to normalize context influence by using the joint self-entropy. Moreover, we note that our work is only focused on measuring the privacy leakage of an augmented context when releasing an LM generation. Context influence is not intended to measure privacy leakage from the parametric knowledge, which is done in memorization works and is orthogonal to our problem setup. 


%% file: body/acknowledgements.tex
\section*{Acknowledgments}
We sincerely thank all reviewers for their time and constructive comments. This material is based upon work supported by NSF award number 2224319 and DGE-1842487, REAL@USC-Meta center, and a VMware gift. The views, opinions, and/or findings expressed are those of the author(s) and should not be interpreted as representing the official views or policies of the U.S. Government.

%% file: body/appendix.tex
\onecolumn
\appendix
\section{Proof of Theorem \ref{thm:priv_loss_pmi}}\label{sec:proof}
We restate the theorem below: 

\begin{theorem}
    Let $\lambda \geq 0$. The context influence of $D_{i, n}$ with the response $y_t$ generated from CID $\overline{p}_\theta$ (Eq. \ref{eq:CID_dist}) is
    \begin{equation}
        \tau_{i, n}(\overline{p}_{\theta, \lambda}, D, \mathbf{x}, \textbf{y}_{<t}, y_t) \propto \lambda \left | \text{pmi}(p_{\theta}(y_t;D,\mathbf{x},\mathbf{y}_{<t})) - \text{pmi}(p_{\theta}(y_t;D \setminus D_{i, n},\mathbf{x},\mathbf{y}_{<t})) \right|. 
    \end{equation}
\end{theorem}

\textit{Proof.} Note that using the definition of CAD (Eq. \ref{eq:cad}), we can write 
\begin{equation}\label{eq:prop_cid}
    \overline{p}_{\theta, \lambda} \propto p_{\theta}(y_t | \mathbf{x}, \mathbf{y}_{<t}) \exp\left[\text{pmi}(p_{\theta}(y_t;D,\mathbf{x},\mathbf{y}_{<t}))\right]^{\lambda}.
\end{equation}

Hence we have 

\begin{align}
    \tau_{i, n}(\overline{p}_{\theta}, D, \mathbf{x}, \textbf{y}_{<t}, y_t) &= \left | \log\left(\overline{p}_{\theta, \lambda}(y_t|D,\mathbf{x}_t,\mathbf{y}_{<t} \right) - \log\left(\overline{p}_{\theta, \lambda}(y_t|D \setminus D_{i, n},\mathbf{x}_t,\mathbf{y}_{<t} \right) \right | \nonumber \\
    &\propto \biggl | \log \left (p_{\theta}(y_t | \mathbf{x}, \mathbf{y}_{<t}) \exp\left[\text{pmi}(p_{\theta}(y_t;D,\mathbf{x},\mathbf{y}_{<t}))\right]^{\lambda} \right ) \nonumber \\
    &- \log \left (p_{\theta}(y_t | \mathbf{x}, \mathbf{y}_{<t}) \exp\left[\text{pmi}(p_{\theta}(y_t;D\setminus D_{i, n},\mathbf{x},\mathbf{y}_{<t}))\right]^{\lambda} \right ) \biggr | \label{eq:use_prop_cid}\\
    &= \left | \lambda \text{pmi}(p_{\theta}(y_t;D,\mathbf{x},\mathbf{y}_{<t})) - \lambda \text{pmi}(p_{\theta}(y_t;D \setminus D_{i, n},\mathbf{x},\mathbf{y}_{<t})) \right | \label{eq:log_prop}\\
    &= \lambda \left | \text{pmi}(p_{\theta}(y_t;D,\mathbf{x},\mathbf{y}_{<t})) - \text{pmi}(p_{\theta}(y_t;D \setminus D_{i, n},\mathbf{x},\mathbf{y}_{<t})) \right| \nonumber
\end{align}

where the proportionality (Eq. \ref{eq:use_prop_cid}) uses the Eq. \ref{eq:prop_cid}, and Eq. \ref{eq:log_prop} uses the product and power rule of logarithms to simplify the expression.  \qed

\section{Proof of CID satisfying $\epsilon$-DP}\label{sec:add_dp}
We will now show how CID can satisfy $n$-gram level $\epsilon$-DP (Definition \ref{def:dp}). First, we are going to slightly modify CID by first selecting $\lambda$ so that we bound the amount of information leaked from a context $D$ when releasing the next token $y_t$. The algorithm can be found in Algorithm \ref{alg:projection}, which follows from \cite{husain2020local, flemings2024differentially}. 

\begin{algorithm}[h]
    \caption{Bounded CID} 
    \label{alg:projection}
    \begin{algorithmic}[1]
        \Function{$\mathcal{P}$}{$p_\theta, D, \mathbf{x}, \mathbf{y}_{<t}, y_t, \epsilon$}
        \State Choose $\lambda \in [0, \infty)$ such that $\left | \log \left (\frac{\overline{p}_\theta(y_t|D, \mathbf{x}, \mathbf{y}_{<t})}{p_\theta(y_t | \mathbf{x}, \mathbf{y}_{<t})} \right ) \right | \leq \frac{\epsilon}{2}$\label{eq:lambda_D}
        \State $\overline{p}_{\theta}(y_t|D,\mathbf{x},\mathbf{y}_{<t})=\text{softmax}[\lambda\text{logit}_\theta(y_t|D,\mathbf{x},\mathbf{y}_{<t})+(1-\lambda)\text{logit}_\theta(y_t|\mathbf{x},\mathbf{y}_{<t})]$
        \State \Return $\overline{p}_{\theta}(y_t|D,\mathbf{x},\mathbf{y}_{<t})$
        \EndFunction
    \end{algorithmic}
\end{algorithm}

Before proving that Algorithm \ref{alg:projection} is $\epsilon$-DP, we introduce a new term to help with the proof. We consider the privacy loss random variable, which is the log probability ratio as a random variable. Drawing $t\sim \mathcal{A}(D)$, we get
\begin{equation}
    \mathcal{L}_{D, D\setminus \{d_i\}} = \log\left(\frac{\Pr[\mathcal{A}(D)=t]}{\Pr[\mathcal{A}(D\setminus\{d_i\}=t]} \right).
\end{equation}
It is immediate from the definition of pure differential privacy (Definition \ref{def:dp}) that $\epsilon$-DP corresponds to $|\mathcal{L}_{D, D\setminus \{d_i\}}|$ being bounded by $\epsilon$ for all neighboring datasets $D, D\setminus \{d_i\}$. Hence, we need to show that the privacy loss random variable of Algorithm \ref{alg:projection} is bounded by $\epsilon$ for neighboring datasets $D, D\setminus D_{i, n}$.

\begin{theorem}
    Let $y_t \sim \mathcal{P}(p_\theta, D, \mathbf{x}, \mathbf{y}_{<t}, y_t, \epsilon)$ be a token generated by the bounded CID from Algorithm \ref{alg:projection}. Then $y_t$ is $\epsilon$-DP with respect to $D$.
\end{theorem}

\textit{Proof.} Let $D$ be a dataset and $D_{i, n}$ be the $i$-th token $n$-gram of $D$. Then for any $y_t \in \mathcal{V}$ where $\mathcal{V}$ is the vocabulary of the LLM $p_{\theta}$, we get the following: 

\begin{align}
    &\left|\log\left(\frac{y_t \sim \mathcal{P}(p_\theta, D, \mathbf{x}, \mathbf{y}_{<t}, y_t, \epsilon)}{y_t \sim \mathcal{P}(p_\theta, D\setminus D_{i, n}, \mathbf{x}, \mathbf{y}_{<t}, y_t, \epsilon)}\right)\right|\nonumber \\ 
    &= \left| \log\left(\frac{\overline{p}_\theta(y_t|D, \mathbf{x}, \mathbf{y}_{<t})}{\overline{p}_\theta(y_t | D\setminus D_{i, n}, \mathbf{x}, \mathbf{y}_{<t})}\right) \right | \nonumber \\
    &= \left|\log\left(\frac{\overline{p}_\theta(y_t|D, \mathbf{x}, \mathbf{y}_{<t}) p_\theta(y_t |\mathbf{x}, \mathbf{y}_{<t})}{\overline{p}_\theta(y_t | D\setminus D_{i, n}, \mathbf{x}, \mathbf{y}_{<t}) p_\theta(y_t | \mathbf{x}, \mathbf{y}_{<t})}\right) \right| \nonumber \\
    &= \left | \log\left(\frac{\overline{p}_\theta(y_t|D, \mathbf{x}, \mathbf{y}_{<t})}{p_\theta(y_t | \mathbf{x}, \mathbf{y}_{<t})}\right) + \log\left(\frac{p_\theta(y_t |\mathbf{x}, \mathbf{y}_{<t})}{\overline{p}_\theta(y_t | D\setminus D_{i, n}, \mathbf{x}, \mathbf{y}_{<t})}\right) \right| \nonumber \\
    &\leq \left | \log\left(\frac{\overline{p}_\theta(y_t|D, \mathbf{x}, \mathbf{y}_{<t})}{p_\theta(y_t | \mathbf{x}, \mathbf{y}_{<t})}\right)\right| + \left | \log\left(\frac{p_\theta(y_t |\mathbf{x}, \mathbf{y}_{<t})}{\overline{p}_\theta(y_t | D\setminus D_{i, n}, \mathbf{x}, \mathbf{y}_{<t})}\right) \right|\label{eq:tri_ineq} \\
    &\leq  \frac{\epsilon}{2} +  \frac{\epsilon}{2} =  \epsilon \label{eq:lambda_def}.
\end{align}

Eq. \ref{eq:tri_ineq} is due to the triangle inequality and Eq. \ref{eq:lambda_def} is from line \ref{eq:lambda_D} from Algorithm \ref{alg:projection}. \qed

\section{Additional Experimental Setup}\label{sec:exam_promp}
\begin{figure}[h]
    \centering
    \begin{subfigure}{0.49\textwidth}
        \begin{mdframed}[backgroundcolor=gray!5, linecolor=black, linewidth=1pt]
        \begin{center}
            \textbf{PubMedQA} 
        \end{center}
        
        \noindent \textcolor{red}{Document: Programmed cell death (PCD) is the regulated death of cells within an organism. The lace plant (Aponogeton madagascariensis) produces perforations in its leaves through ...} 
        \\
        
        \noindent \textcolor{blue}{Do mitochondria play a role in remodelling lace plant leaves during programmed cell death?}
        \end{mdframed}
    \end{subfigure}
    \begin{subfigure}{0.49\textwidth}
        \begin{mdframed}[backgroundcolor=gray!5, linecolor=black, linewidth=1pt]
        \begin{center}
            \textbf{CNN} 
        \end{center}
        
        \noindent \textcolor{red}{News article: (CNN)The Palestinian Authority officially became the 123rd member of the International Criminal Court on Wednesday, a step that gives the court jurisdiction over alleged crimes ...} 
        \\
        
        \noindent \textcolor{blue}{Summary of the above news article:}
        \\
        \end{mdframed}
    \end{subfigure}
    \caption{Example prompts with context used for PubMedQA and CNN where \textcolor{red}{red text} is the context $\boldsymbol{D}$ and \textcolor{blue}{blue text} is the query $\boldsymbol{x}$.}
    \label{fig:exam_prompts}
\end{figure}

\begin{figure}[h]
    \centering
    \begin{subfigure}{0.49\textwidth}
        \begin{mdframed}[backgroundcolor=gray!5, linecolor=black, linewidth=1pt]
        \begin{center}
            \textbf{PubMedQA} 
        \end{center}
        
        \noindent \textcolor{red}{Document: .} 
        \\
        
        \noindent \textcolor{blue}{Do mitochondria play a role in remodelling lace plant leaves during programmed cell death?}
        \end{mdframed}
    \end{subfigure}
    \begin{subfigure}{0.49\textwidth}
        \begin{mdframed}[backgroundcolor=gray!5, linecolor=black, linewidth=1pt]
        \begin{center}
            \textbf{CNN} 
        \end{center}
        
        \noindent \textcolor{red}{News article: .} 
        \\
        
        \noindent \textcolor{blue}{Summary of the above news article:}
        \\\\
        \end{mdframed}
    \end{subfigure}
    \caption{Example prompts without context used for PubMedQA and CNN where \textcolor{red}{red text} is the context $\boldsymbol{D}$ and \textcolor{blue}{blue text} is the query $\boldsymbol{x}$.}
    \label{fig:exam_no_contextprompts}
\end{figure}

Figures \ref{fig:exam_prompts} and \ref{fig:exam_no_contextprompts} illustrate exemplar prompts with and without context used for each dataset in our experiments for Sections \ref{sec:input_regurg} and \ref{sec:factors}.

For hardware, all of our experiments used one A100 40GB GPU, except for the experiments using 30B and 66B which used two and three A100 40GB GPUs, respectively. Each experiment usually takes around 15 mins to run for OPT-1.3B, except for the $i$-th token $n$-gram experiments (Section \ref{sec:ith_n-gram} which take longer depending on how small the token $n$-gram is. For software, our summarization quality evaluation is based on the code from \citet{xu2023context}, which is freely available on GitHub \footnote{\url{https://github.com/zhichaoxu-shufe/context-aware-decoding-qfs}}. All datasets and models used in our experiments are freely available at Hugging Face, and our research does not conflict with their intended use cases, which is to evaluate text generation quality and privacy. The CNN-DM dataset follows the apache-2.0 license, LLaMA 3 follows the Llama 3.2 Community License Agreement, which we agreed to before evaluating, and GPT-Neo follows the MIT license, all of which we ensured not to go against.

\section{Additional Experimental Results}\label{sec:add_exp_results}
\input{figures/analysis_temp}

\input{figures/analysis_lambda}

Figure \ref{fig:abal_temp} shows the average context influence, ROUGE-L, and FactKB across different temperature values. We observe that as $\tau$ approaches zero, the model is influenced by the context exponentially, with moderate improvements in similarity. This is because as $\tau$ approaches zero, the decoding becomes equivalent to argmax, where the token with the highest probability is selected. Hence, there is less entropy in the decoding since the output distributions are sharper, so there is more divergence between the posterior and prior distributions (larger PMI). However, the faithfulness actually decreases once $\tau < 0.4$, demonstrating that less randomness during decoding can result in generations that are not as faithful to the context.

Figure \ref{fig:abal_lambda} shows the average context influence, ROUGE-L, and FactKB across different context influence levels $\lambda$. Our results suggest that a higher average influence of the context leads to more faithfulness to the context (higher FactKB), but for $\lambda>1.25$, the similarity of the generated response to the gold response slightly degrades.

Next, we qualitatively analyze generations from LLaMA-3 (8B) for CNN-DM in Table \ref{tbl:qualitative}. We observed that many of the $\lambda=1.5$ generations are regurgitating the context, highlighting that amplifying the context increases surfacing of contextual information. Regular decoding, $\lambda=1.0$, is also prone to regurgitating contextual information but is not as severe. In particular, both $\lambda=1.5$ and $\lambda=1.0$ contain "UFO" in their generations, information likely derived verbatim from the context. On the other hand, $\lambda=0.5$ does not contain UFO and instead contains "flying vehicle," which is broadly relevant but does not appear verbatim in the context, indicating a strong reliance on parametric knowledge. 

\input{figures/qualitative_analysis}

%% file: figures/analysis_temp.tex
\begin{figure}[h]
    \centering
    \begin{subfigure}{0.32\linewidth}
    \centering
        \begin{tikzpicture}
            \begin{axis}[
                outer sep=0pt,
                inner sep=0pt,
                width=1.2\linewidth,
                height=5cm,
                xlabel={temperature $T$},
                ylabel={$\mathbb{E}[f_{\text{infl}}(\overline{p}_{\theta})]$} ,
                xmin=0,
                xmax=1.0,
                ymin=30,
                ymax=190,
                ytick={70, 110, 150, 190},
                xtick={0, 0.25, 0.5, 0.75, 1.0},
                label style={inner sep=0pt, outer sep=0pt, font=\tiny},
                tick label style={inner sep=1pt, font=\tiny},
                xmajorgrids=true,
                ymajorgrids=true,
            ]
            \addplot[
                color=blue,
                mark=*,
            ]
            coordinates {
            (0.2, 189.12)
            (0.4, 95.39)
            (0.6, 61.07)
            (0.8, 45.66)
            (1.0, 32.67)
            };
            \end{axis}
        \end{tikzpicture} 
    \end{subfigure}
    \hfill 
    \begin{subfigure}{0.32\linewidth}
    \centering
        \begin{tikzpicture}
            \begin{axis}[
                outer sep=0pt, 
                inner sep=0pt,
                outer sep=0pt,
                width=1.2\linewidth,
                height=5cm,
                xlabel={temperature $T$},
                ylabel={ROUGE-L} ,
                xmin=0,
                xmax=1.0,
                ymin=13,
                ymax=21,
                ytick={15, 17, 19, 21},
                xtick={0, 0.25, 0.5, 0.75, 1.0},
                label style={inner sep=0pt, font=\tiny},
                tick label style={inner sep=1pt, font=\tiny},
                xmajorgrids=true,
                ymajorgrids=true,
            ]
            \addplot[
                color=blue,
                mark=*,
            ]
            coordinates {
            (0.2, 19.77)
            (0.4, 19.39)
            (0.6, 18.22)
            (0.8, 16.51)
            (1.0, 13.56)
            };
            \end{axis}
        \end{tikzpicture} 
    \end{subfigure}
    \hfill 
    \begin{subfigure}{0.32\linewidth}
    \centering
        \begin{tikzpicture}
            \begin{axis}[
                outer sep=0pt,
                inner sep=0pt, 
                width=1.2\linewidth,
                height=5cm,
                xlabel={temperature $T$},
                ylabel={FactKB} ,
                xmin=0.0,
                xmax=1.0,
                ymin=18,
                ymax=46,
                ytick={25, 32, 39, 46},
                xtick={0.0, 0.25, 0.5, 0.75, 1.0},
                label style={inner sep=0pt, font=\tiny},
                tick label style={inner sep=1pt, font=\tiny},
                xmajorgrids=true,
                ymajorgrids=true,
            ]
            \addplot[
                color=blue,
                mark=*,
            ]
            coordinates {
            (0.2, 41.30)
            (0.4, 44.81)
            (0.6, 44.22)
            (0.8, 37.38)
            (1.0, 22.34)
            };
            \end{axis}
        \end{tikzpicture} 
    \end{subfigure}
    \caption{Measuring context influence, ROUGE-L, and FactKB with respect to different temperature $\tau$ values on PubMedQA for OPT-6.7B on PubMedQA using $\lambda=1.0$}
    \label{fig:abal_temp}
\end{figure}

%% file: figures/analysis_lambda.tex
\begin{figure}[!h]
    \centering
    \begin{subfigure}{0.32\linewidth}
    \centering
        \begin{tikzpicture}
            \begin{axis}[
                outer sep=0pt,
                inner sep=0pt,
                width=1.2\linewidth,
                height=5cm,
                xlabel={$\lambda$},
                ylabel={$\mathbb{E}[f_{\text{infl}}(\overline{p}_{\theta})]$} ,
                xmin=0,
                xmax=2.0,
                ymin=0,
                ymax=150,
                ytick={25, 50, 100, 150},
                xtick={0.0, 0.5, 1.0, 1.5, 2.0},
                label style={inner sep=0pt, outer sep=0pt, font=\tiny},
                tick label style={inner sep=1pt, font=\tiny},
                xmajorgrids=true,
                ymajorgrids=true,
            ]
            \addplot[
                color=blue,
                mark=*,
            ]
            coordinates {
            (0.25, 6.30)
            (0.5, 13.40)
            (0.75, 24.57)
            (1.0, 40.69)
            (1.25, 70.52)
            (1.5, 97.95)
            (2.0, 146.67)
            };
            \end{axis}
        \end{tikzpicture} 
    \end{subfigure}
    \hfill 
    \begin{subfigure}{0.32\linewidth}
    \centering
        \begin{tikzpicture}
            \begin{axis}[
                outer sep=0pt, 
                inner sep=0pt,
                outer sep=0pt,
                width=1.2\linewidth,
                height=5cm,
                xlabel={$\lambda$},
                ylabel={ROUGE-L} ,
                xmin=0,
                xmax=2.0,
                ymin=15,
                ymax=17.1,
                ytick={15.5, 16, 16.5, 17.0},
                xtick={0.0, 0.5, 1.0, 1.5, 2.0},
                label style={inner sep=0pt, font=\tiny},
                tick label style={inner sep=1pt, font=\tiny},
                xmajorgrids=true,
                ymajorgrids=true,
            ]
            \addplot[
                color=blue,
                mark=*,
            ]
            coordinates {
            (0.25, 15.16)
            (0.5, 15.41)
            (0.75, 16.18)
            (1.0, 16.51)
            (1.25, 17.02)
            (1.5, 16.96)
            (2.0, 16.92)
            };
            \end{axis}
        \end{tikzpicture} 
    \end{subfigure}
    \hfill 
    \begin{subfigure}{0.32\linewidth}
    \centering
        \begin{tikzpicture}
            \begin{axis}[
                outer sep=0pt,
                inner sep=0pt, 
                width=1.2\linewidth,
                height=5cm,
                xlabel={$\lambda$},
                ylabel={FactKB} ,
                xmin=0.0,
                xmax=2.0,
                ymin=25,
                ymax=55,
                ytick={35, 45, 55},
                xtick={0.0, 0.5, 1.0, 1.5, 2.0},
                label style={inner sep=0pt, font=\tiny},
                tick label style={inner sep=1pt, font=\tiny},
                xmajorgrids=true,
                ymajorgrids=true,
            ]
            \addplot[
                color=blue,
                mark=*,
            ]
            coordinates {
            (0.25, 29.74)
            (0.5, 31.40)
            (0.75, 33.74)
            (1.0, 38.78)
            (1.25, 44.24)
            (1.5, 48.81)
            (2.0, 54.08)
            };
            \end{axis}
        \end{tikzpicture} 
    \end{subfigure}
    \caption{Measuring context influence, ROUGE-L, and FactKB with respect to different $\lambda$ values on PubMedQA for OPT-1.3B.}
    \label{fig:abal_lambda}
\end{figure}

%% file: figures/qualitative_analysis.tex
\begin{table*}[!ht]
\small
\begin{tabularx}{\textwidth}{lX}
\toprule
\multicolumn{2}{c}{\textbf{CNN-DM}} \\
\midrule
Article & ... Luckily, Japanese can sleep soundly in their beds tonight as the government's top military official earnestly revealed that the country's Air Self Defense Force (ASDF) had never encountered an \colorbox{red}{extraterrestrial unidentified flying object}. Responding to a query from flamboyant former wrestler-turned-lawmaker Antonio Inoki, Defense Minister Gen Nakatani told the Diet, Japan's parliament, that his jets had, to date, never come across any \colorbox{red}{UFOs} from outer space. ... \\[1.75cm]

\multirowcell{2}{$\lambda=1.5$} &  Japanese can sleep soundly in their beds tonight as the government's top military official earnestly revealed that the country's Air Self Defense Force (ASDF) had never encountered an \colorbox{red}{extraterrestrial unidentified flying object}. \\[0.6cm]

\multirowcell{3}{$\lambda=1$} &  in a interview with Japanese defense minister, politician Antonio Inoki asked the defense minister about aliens and \colorbox{red}{UFOs} and the defense minister answered that the Air Self Defense Force (ASDF) has never encountered one.  \\ [0.6cm]

$\lambda=0.5$ &  The article discusses the topic of the possible appearance of aliens and their \colorbox{yellow}{flying vehicles} in the skies over Japan. The author of the article recalls that recently there was a flight of a mysterious object in the sky over Japan. \\
\bottomrule
\end{tabularx}
\caption{Qualitative examples from LLaMA 3 using different influence levels of CID.}
\label{tbl:qualitative}
\end{table*}